\documentclass[runningheads]{llncs}
\usepackage[T1]{fontenc}
\usepackage{graphicx}
\usepackage{amsmath,amssymb}
\usepackage{booktabs,tabularx}
\usepackage{microtype}
\usepackage{flafter}
\usepackage[section]{placeins}
\usepackage{xurl}
\usepackage[hidelinks]{hyperref}
\graphicspath{{figures/}}
\begin{document}
\title{Radiomics-Conditioned Modulation of RenalCLIP Features for Clear Cell Renal Cell Carcinoma Classification}
\titlerunning{Radiomics-Conditioned RenalCLIP Feature Modulation}
\author{
Yuan Liang\inst{1,3}\textsuperscript{*} \and
Sourav Bhattacharjee\inst{2} \and
Abraham Campbell\inst{3}
}
\authorrunning{Y. Liang et al.}
\institute{
Research Ireland Centre for Research Training in Machine Learning \and
School of Veterinary Medicine, University College Dublin, Dublin, Ireland \and
School of Computer Science, University College Dublin, Dublin, Ireland \\
\textsuperscript{*}Corresponding author:
\email{yuan.liang@ucdconnect.ie}
}
\maketitle

\begin{abstract}
Radiomics provides quantitative descriptions of tumour appearance that may complement disease-specific foundation models in small labelled cohorts. We investigate this complementarity for computed tomography-based classification of clear cell renal cell carcinoma. Our framework uses radiomics to modulate RenalCLIP features through feature-wise linear modulation (FiLM), while retaining a direct radiomics contribution. Internal testing and external validation compare it with conventional fusion strategies and reference classifiers. The FiLM model achieves an area under the receiver operating characteristic curve (AUC) of 0.804 internally and 0.854 externally, with the highest mean AUC among the evaluated RenalCLIP fusion strategies in both cohorts. Pathway ablations examine the contributions of conditional modulation and the direct radiomics residual, while feature permutation highlights the role of tumour texture. These findings support radiomics as a useful complement to RenalCLIP in a small labelled cohort and identify FiLM as an effective approach to integrating their representations for robust renal tumour classification.
\keywords{Renal cell carcinoma \and Radiomics \and Foundation models \and Feature fusion \and Feature-wise linear modulation}
\end{abstract}

\section{Introduction}

Preoperative characterisation of renal masses requires distinguishing tumours with overlapping imaging appearances. Clear cell renal cell carcinoma (ccRCC) can exhibit enhancement patterns different from those of other renal tumours, and multiphasic computed tomography (CT) measurements can assist this distinction~\cite{young2013clear}. Nevertheless, a lesion contains spatial variation in intensity and morphology that is only partly represented by conventional measurements. Computational analysis offers a means of incorporating this information into tumour classification.

Radiomics characterises a segmented tumour through quantitative measurements of shape, intensity, and texture. These descriptors summarise three-dimensional geometry, intensity distributions, and spatial relationships between grey levels. They capture patterns that can be difficult to quantify consistently through visual inspection and have been investigated as markers of tumour phenotype~\cite{aerts2014decoding,lambin2017radiomics}. Their explicit definitions link classifier inputs to measurable tumour properties.

This structured representation is particularly relevant when labelled data are limited. Its feature extractor does not require target-cohort training, and feature selection can prioritise reproducible, nonredundant measurements. Although predefined descriptors cannot encompass every aspect of tumour appearance, they provide a compact quantitative basis that may complement learned image representations~\cite{lambin2017radiomics}.

Deep learning provides another way to represent the same image. Residual networks learn spatial filters from data~\cite{he2016resnet}, while vision--language pretraining relates image appearance to associated textual information~\cite{radford2021clip}. RenalCLIP applies disease-specific visual--textual learning to kidney cancer~\cite{tao2026renalclip}. Its image representation incorporates information learned from a large pretraining cohort, but transfer to a particular histological distinction still depends on how the representation is adapted. We hypothesise that explicit radiomic descriptors can supplement RenalCLIP features when the target task has relatively few labelled examples.

The way these representations interact may influence how effectively that supplementary information is used. Concatenation makes both vectors available to a common classifier. Gated fusion learns their relative weighting~\cite{arevalo2017gmu}. Feature-wise linear modulation (FiLM) instead uses a conditioning input to generate an affine transformation of another representation~\cite{perez2018film}. Radiomics can therefore do more than contribute additional classifier inputs. It can guide how image features are scaled and shifted for each patient.

We propose a radiomics-conditioned residual FiLM framework for ccRCC classification. The model uses radiomics both to modulate RenalCLIP features and to provide a direct residual representation for the classifier. We compare this formulation with concatenation, alternative gates, and globally shared affine modulation, together with conventional radiomics and convolutional references. Internal and external evaluation assesses classification performance, while pathway ablations and radiomics permutation analysis examine how the model uses the additional quantitative information.

\section{Methods}

\subsection{Study Cohorts}

The internal cohort comprised 396 KiTS23 patients~\cite{heller2023kits23} with CT, tumour masks, histological labels, and complete feature records. ccRCC was defined as the positive class and the remaining histologies as the negative class. The fixed patient-level split comprised 237 training, 99 validation, and 60 test patients, with stratification used to maintain an approximately 70\% ccRCC prevalence across the three subsets. Multiple tumour delineations were combined using STAPLE consensus where available~\cite{warfield2004staple}.

The external cohort combined RCC-AID data~\cite{de2026rcc} with previously assembled TCGA cases using AIMI annotations~\cite{vanoss2023aimi}. After patient-level deduplication and image-quality review, 111 patients remained, comprising 85 ccRCC and 26 non-ccRCC cases (76.6\% ccRCC). All methods were evaluated on the same selected CT scan and tumour mask for each patient.

\subsection{CT Processing and Feature Representations}

\subsubsection{Radiomics.}
CT volumes were consistently oriented, resampled to 1-mm isotropic spacing, and clipped to $[-150,200]$ Hounsfield units (HU). Masks underwent nearest-neighbour resampling. Features were extracted with PyRadiomics~\cite{vangriethuysen2017radiomics}, using definitions described by the Image Biomarker Standardization Initiative~\cite{zwanenburg2020ibsi}. Original-image shape descriptors were combined with first-order and texture features from the original and Laplacian-of-Gaussian (LoG) images. LoG scales were $\sigma\in\{1,2,3\}$~mm and the intensity bin width was 25~HU. Filtering preceded cropping.

The initial 386 descriptors were reduced to 73 using inter-annotator ICC(3,1) $\geq0.75$~\cite{shrout1979icc}, exclusion below the 20th percentile of median absolute deviation (MAD), and correlation filtering at $|\rho|>0.95$. Average-linkage clustering with distance $1-|\rho|$ and a cut of 0.2 retained the highest-MAD feature per cluster. Median imputation and standardisation parameters were fitted on training patients and applied unchanged to subsequent datasets, producing $\mathbf{x}\in\mathbb{R}^{73}$.

\subsubsection{RenalCLIP.}

The public RenalCLIP image encoder~\cite{tao2026renalclip} provided disease-specific image representations. Its pretraining data came from independent institutional cohorts in China and did not overlap with the present study datasets. The 3D ResNet18 encoder and its batch-normalisation statistics were held fixed. We pooled its third residual-stage feature map using tumour-mask weights, producing $\mathbf{z}\in\mathbb{R}^{256}$ after L2 normalisation. Mask voxels were distributed trilinearly to stride-aligned centres on the $2\times8\times8$ feature grid. The pooled vector was the weighted mean of these features. This emphasised tumour-associated locations while retaining contextual receptive fields.

An $87.5\times87.5\times100$~mm field of view was centred on the tumour centroid in the axial slice with the largest tumour area. Trilinear sampling produced $140\times140\times32$ voxels, followed by a $128\times128\times32$ centre crop, giving a final $80\times80\times100$~mm field of view. Windowing at centre 50 HU and width 500 HU mapped intensities to [0,1]. The same processing was applied across cohorts.

\subsection{Radiomics-Conditioned Residual Modulation}

\begin{figure}[!htbp]
\centering
\includegraphics[width=\textwidth]{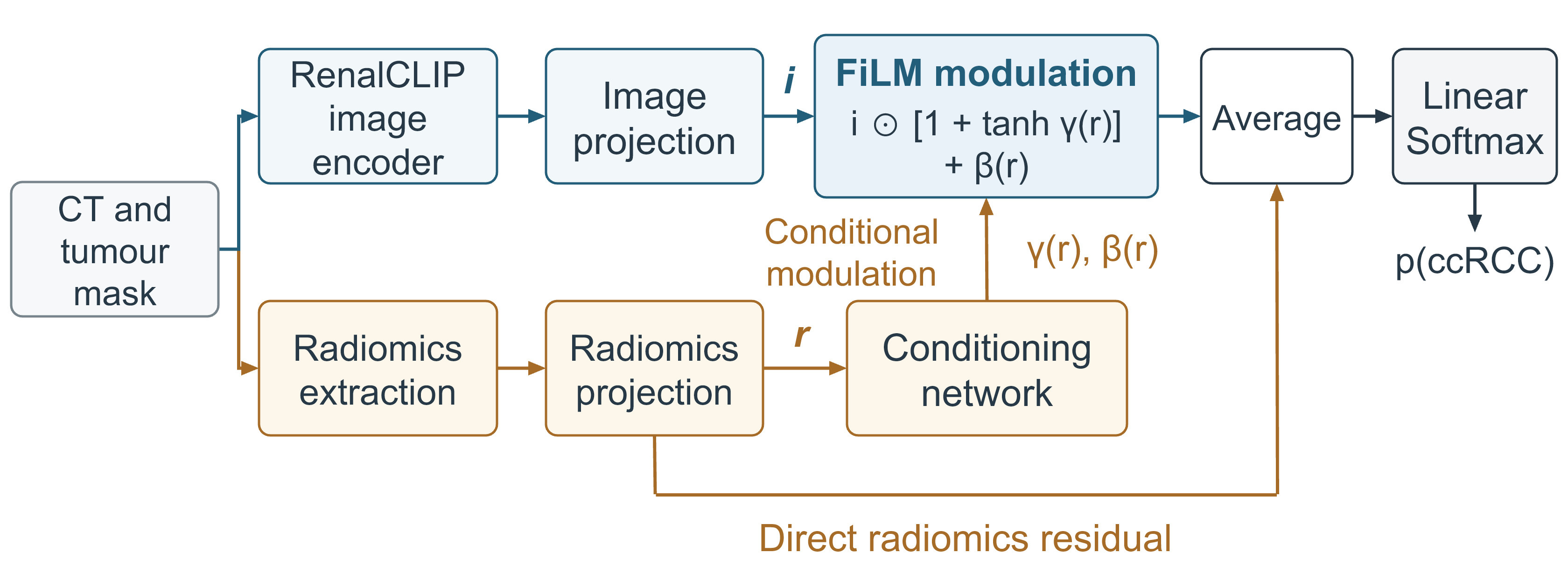}
\caption{Radiomics-conditioned residual FiLM. Radiomics guides the affine transformation of image features and contributes directly to the fused representation. The image encoder is fixed while the projections, conditioning network, and classifier are learned.}
\label{fig:architecture}
\end{figure}

The proposed framework uses radiomics to condition the image representation while preserving a direct radiomics contribution (Fig.~\ref{fig:architecture}). Let $P_I$ and $P_R$ denote independently learned projections with ReLU activation and dropout. They map image and radiomics inputs into a common feature space,
\begin{equation}
\mathbf{i}=P_I(\mathbf{z}),\qquad
\mathbf{r}=P_R(\mathbf{x}),\qquad
\mathbf{i},\mathbf{r}\in\mathbb{R}^{128}.
\label{eq:projections}
\end{equation}
A conditioning network with dimensions $128\rightarrow128\rightarrow256$ generates the patient-specific modulation parameters,
\begin{equation}
\operatorname{concat}\!\left(\boldsymbol{\gamma}(\mathbf{r}),\boldsymbol{\beta}(\mathbf{r})\right)
=W_2\operatorname{ReLU}(W_1\mathbf{r}+\mathbf{b}_1)+\mathbf{b}_2.
\label{eq:conditioning}
\end{equation}
The fused representation is defined as
\begin{equation}
\mathbf{h}=\frac{1}{2}\left[
\mathbf{i}\odot\{1+\tanh\boldsymbol{\gamma}(\mathbf{r})\}
+\boldsymbol{\beta}(\mathbf{r})+\mathbf{r}\right],
\label{eq:film}
\end{equation}
where $\odot$ denotes element-wise multiplication. The multiplicative term adapts the contribution of each image coordinate within a bounded range. The additive term adjusts its offset, and the direct residual preserves the projected radiomics information. Zero initialisation of the final conditioning layer gives $\mathbf{h}=(\mathbf{i}+\mathbf{r})/2$ at the start of optimisation. A linear classifier followed by softmax maps $\mathbf{h}$ to the ccRCC probability.

\subsection{Comparison Methods}

A radiomics-based radial-basis-function support-vector classifier (SVC)~\cite{cortes1995svm} served as the conventional reference. Convolutional comparisons comprised a 3D ResNet18 trained from scratch, together with its radiomics-gated and radiomics-FiLM variants. These models used normalised $128^3$ tumour-centred CT inputs and 512-dimensional ResNet18 representations for fusion.

For RenalCLIP, a multilayer perceptron with dimensions $256\rightarrow128\rightarrow2$ classified image features. For concatenation, the 256-dimensional RenalCLIP representation and the 73-dimensional radiomics vector were concatenated directly, yielding a 329-dimensional input to a $329\rightarrow64\rightarrow2$ classifier. Gated fusion combined the projections as
\begin{equation}
\mathbf{h}_{g}=\mathbf{g}\odot\mathbf{i}+(1-\mathbf{g})\odot\mathbf{r}.
\label{eq:gate}
\end{equation}
The gate was conditioned either on radiomics alone or jointly on both representations. In the first case, $\mathbf{g}=\operatorname{sigmoid}(G_R(\mathbf{r}))$. In the second, $G_J$ received the concatenated image and radiomics projections. Both generators had a 128-dimensional hidden layer. A global affine comparison replaced the patient-specific $\boldsymbol{\gamma}$ and $\boldsymbol{\beta}$ in Eq.~\ref{eq:film} with vectors shared across patients. This comparison assesses the role of adaptive conditioning within the same residual formulation.

\subsection{Training and Evaluation}

RenalCLIP-based classifiers were optimised with cross-entropy and AdamW at learning rate $10^{-3}$, with batch size 16 and zero weight decay. Dropout of 0.2 was used in the 128-dimensional projections and image-only classifier. The concatenation classifier used no dropout. Convolutional references used class-weighted cross-entropy and Adam at learning rate $10^{-4}$ with mixed precision. Their training augmentation included spatial flips, rotations, and affine perturbations. Models were selected according to internal validation AUC and applied to the test and external cohorts without further fitting.

The primary metric was AUC, with average precision (AP) as a complementary measure. RenalCLIP-based methods were evaluated over three independent training runs. The SVC and convolutional references each used one trained model. Internal results report mean and sample standard deviation (SD) for repeated runs and 95\% confidence intervals (CIs) for single-model references. External results report patient-bootstrap CIs. Each bootstrap sample preserved class counts and the correspondence of predictions across runs. For methods with repeated training, metrics were calculated separately and then averaged. CIs used 2,000 resamples~\cite{efron1979bootstrap} and quantify patient-sampling uncertainty for the fitted models.

\subsection{Pathway Ablations and Feature Permutation}

Two inference-time ablations examined the radiomics pathways in the fitted FiLM model. Removing the direct residual gives $\mathbf{h}_{\mathrm{no\ direct}}=\mathbf{h}-\mathbf{r}/2$, preserving conditional modulation. Disabling modulation gives $\mathbf{h}_{\mathrm{no\ modulation}}=(\mathbf{i}+\mathbf{r})/2$, preserving the projected representations. All other weights were held fixed.

For feature importance, each radiomics descriptor was independently permuted across internal-test patients 50 times~\cite{fisher2019reliance}. The same permutations were used across training runs. The radiomics projection and modulation parameters were recomputed after each permutation. Importance was defined as the decrease in AUC, averaged first across permutations and then across runs.

\section{Results and Discussion}

\subsection{Internal Classification Performance}

FiLM achieved the highest mean internal AUC among the evaluated RenalCLIP fusion strategies, reaching 0.804 (Table ~\ref{tab:internal}) and improving on RenalCLIP alone by approximately 9.8 percentage points. Global affine modulation achieved the next highest mean AUC, followed by the gated and concatenation approaches. The consistent improvement of the fusion approaches over the image-only classifier supports supplementing transferred features with radiomics. 

\begin{table}[!htbp]
\caption{Internal-test performance. RenalCLIP methods report three-run mean $\pm$ SD. Single-model references report 95\% CIs. All fusion methods include radiomics. Ablations use the fitted FiLM models.}
\label{tab:internal}
\centering\small
\setlength{\tabcolsep}{3pt}
\begin{tabularx}{\textwidth}{@{}Xrr@{}}
\toprule
Method & AUC & AP \\
\midrule
Radiomics SVC & 0.7659 [0.6296, 0.8836] & 0.8938 [0.8259, 0.9500] \\
3D ResNet18 & 0.5324 [0.3651, 0.6971] & 0.7376 [0.6450, 0.8459] \\
ResNet18 + rad. gate & 0.7844 [0.6428, 0.9101] & 0.8651 [0.7736, 0.9615] \\
ResNet18 + rad. FiLM & 0.6865 [0.5384, 0.8307] & 0.8205 [0.7305, 0.9245] \\
\midrule
RenalCLIP only & 0.7063 $\pm$ 0.0026 & 0.7921 $\pm$ 0.0011 \\
\quad + concatenation & 0.7866 $\pm$ 0.0068 & 0.8554 $\pm$ 0.0091 \\
\quad + radiomics gate & 0.7928 $\pm$ 0.0179 & 0.8557 $\pm$ 0.0184 \\
\quad + joint gate & 0.7928 $\pm$ 0.0181 & 0.8640 $\pm$ 0.0080 \\
\quad + global affine & 0.7989 $\pm$ 0.0182 & 0.8596 $\pm$ 0.0145 \\
\textbf{\quad + conditional FiLM} & 0.8038 $\pm$ 0.0081 & 0.8714 $\pm$ 0.0059 \\
\midrule
\multicolumn{3}{l}{\textit{Inference-time pathway ablations of conditional FiLM}} \\
Without direct residual & 0.7959 $\pm$ 0.0038 & 0.8614 $\pm$ 0.0102 \\
Without modulation & 0.8016 $\pm$ 0.0210 & 0.8689 $\pm$ 0.0204 \\
\bottomrule
\end{tabularx}

\end{table}

The similar internal performance of global affine modulation and FiLM indicates that a shared transformation also captured useful relationships between the representations. SVC remained competitive and achieved the highest internal AP, highlighting the discriminatory information already present in radiomics. Within the convolutional comparisons, gating performed better than FiLM. The benefit of a fusion strategy therefore depended on the representation and learning configuration.

\subsection{External Evaluation}

FiLM achieved the highest mean external AUC among the evaluated RenalCLIP fusion strategies, reaching 0.854 (Table~\ref{tab:external}). This exceeded the image-only classifier by 9.86 percentage points. SVC retained a slightly higher AUC, whereas FiLM achieved a higher AP than SVC.

\begin{table}[!htbp]
\caption{External performance on 111 patients. Brackets denote 95\% patient-bootstrap CIs. For methods with three training runs, the point estimate and each bootstrap replicate are means of individual-model metrics.}
\label{tab:external}
\centering\small
\setlength{\tabcolsep}{3pt}
\begin{tabularx}{\textwidth}{@{}Xrr@{}}
\toprule
Method & AUC & AP \\
\midrule
Radiomics SVC & 0.8593 [0.7611, 0.9362] & 0.9378 [0.8796, 0.9807] \\
3D ResNet18 & 0.5853 [0.4566, 0.7088] & 0.8404 [0.7924, 0.8926] \\
ResNet18 + rad. gate & 0.8371 [0.7516, 0.9091] & 0.9479 [0.9190, 0.9723] \\
ResNet18 + rad. FiLM & 0.7620 [0.6629, 0.8498] & 0.9221 [0.8876, 0.9543] \\
\midrule
RenalCLIP only & 0.7549 [0.6422, 0.8537] & 0.9094 [0.8632, 0.9510] \\
\quad + concatenation & 0.8484 [0.7721, 0.9158] & 0.9542 [0.9304, 0.9755] \\
\quad + radiomics gate & 0.8416 [0.7617, 0.9139] & 0.9500 [0.9228, 0.9741] \\
\quad + joint gate & 0.8388 [0.7603, 0.9101] & 0.9491 [0.9228, 0.9724] \\
\quad + global affine & 0.8398 [0.7602, 0.9118] & 0.9489 [0.9212, 0.9730] \\
\textbf{\quad + conditional FiLM} & 0.8535 [0.7781, 0.9244] & 0.9528 [0.9253, 0.9763] \\
\bottomrule
\end{tabularx}

\end{table}

FiLM achieved the highest mean AUC in both the internal and external RenalCLIP fusion comparisons, although the margins over global affine modulation internally and concatenation externally were small. Radiomics provided a consistent improvement across cohorts, with the integration strategy contributing a smaller additional effect. 

\subsection{Pathway Ablations}

Removing the direct radiomics residual and disabling conditional modulation reduced mean internal AUC by 0.79 and 0.22 percentage points, respectively (Table~\ref{tab:internal}). Both pathways contributed numerically to the fitted model, with a larger change after removing the direct residual. Retaining modulation allowed radiomics to influence predictions indirectly, while retaining the residual preserved its explicit contribution.

The effects were modest and indicate overlapping information routes. Additive modulation can itself carry radiomics information, partly compensating for removal of the direct residual. These ablations describe dependence within the fitted model, rather than the performance of independently retrained alternatives. They support using the complete formulation without establishing that either pathway is individually indispensable.

\subsection{Radiomics Feature Importance}

\begin{figure}[!ht]
\centering
\includegraphics[width=\textwidth]{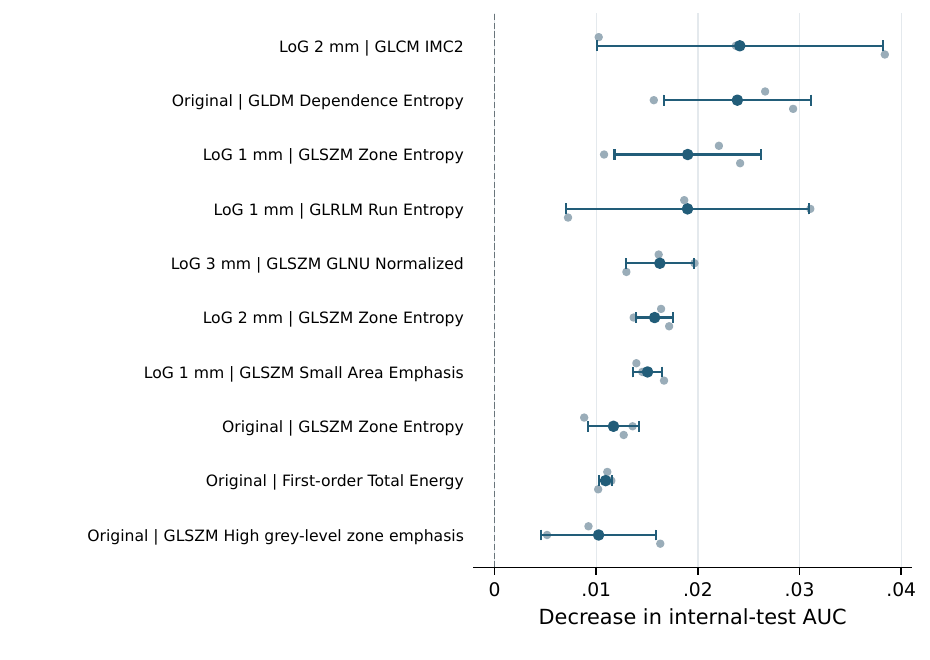}
\caption{The ten radiomics features with the largest mean AUC decrease under permutation. Blue markers and bars show mean $\pm$ SD across runs, and grey markers show individual-run means over 50 permutations. GLCM, GLRLM, GLSZM, and GLDM denote co-occurrence, run-length, size-zone, and dependence matrices. GLNU denotes grey-level nonuniformity.}
\label{fig:importance}
\end{figure}

Texture descriptors dominated the highest-ranked permutation effects (Fig.~\ref{fig:importance}). LoG 2-mm GLCM information measure of correlation 2, original-image GLDM dependence entropy, and LoG 1-mm zone entropy were the leading features. These descriptors quantify different aspects of spatial intensity organisation and heterogeneity. Their prominence indicates that the additional radiomics information used by the model extended beyond tumour size or a single intensity summary.

Because a permuted descriptor affects both the radiomics projection and its modulation parameters, these scores reflect its contribution to the full fusion model. Correlated descriptors may share predictive information, so a small individual permutation effect does not imply that the corresponding image property is unimportant~\cite{fisher2019reliance}. The analysis provides a quantitative account of model reliance, rather than a direct biological attribution.

\subsection{Complementarity and Fusion Design}

The central finding is that explicit radiomic descriptors supplement RenalCLIP for classification with limited labelled data. RenalCLIP reflects patterns learned through disease-specific pretraining, whereas radiomics directly supplies predefined intensity and texture measurements. This distinction provides a plausible basis for complementarity. Information implicit in a pooled image representation may become easier to exploit when an associated quantitative descriptor is supplied explicitly.

FiLM incorporates that information through a transformation conditioned on each patient's radiomics profile. Multiplicative modulation changes the contribution of image coordinates, while additive modulation adjusts their offsets. The direct residual preserves descriptors that are already useful for classification. Together, these mechanisms allow radiomics to contribute both explicit predictive information and context for adapting the learned representation.

FiLM achieved the highest mean AUC within the RenalCLIP fusion comparisons in both cohorts, suggesting that patient-specific conditioning can be useful for integrating the two representations. However, the competitive performance of global affine modulation, concatenation, and gated fusion shows that simpler interactions also exploit the radiomic descriptors effectively. The modest differences between fusion methods, together with the strong SVC results, indicate that radiomics supplies much of the available discriminatory information. The most directly supported complementarity is radiomics supplementing the transferred image representation, while the incremental value of conditional modulation remains modest.

\subsection{Limitations and Future Work}

The study is retrospective and the internal test cohort is relatively small. The convolutional and SVC references use single trained models, while RenalCLIP methods use repeated runs. Comparisons between frozen-feature and end-to-end approaches should be interpreted cautiously because their preprocessing and optimisation procedures differed. Larger independent cohorts and repeated training of all references would provide a more precise assessment of their relative performance. Future work should examine how conditional fusion behaves across acquisition settings and whether its effects persist under changes in segmentation and feature extraction.

\section{Conclusion}

Radiomics provides complementary information for RenalCLIP-based ccRCC classification in a small labelled cohort. Residual FiLM uses these descriptors to modulate image features while preserving their direct contribution. It achieved AUCs of 0.804 internally and 0.854 externally, with the highest mean AUC among the evaluated RenalCLIP fusion strategies in both cohorts. Pathway ablations indicated overlapping radiomics contributions, and permutation analysis highlighted texture descriptors. These findings support conditional modulation for integrating the representations and motivate confirmation across larger, independent cohorts.

% \subsubsection{Acknowledgements.}
% \pending{funding and acknowledgements.}
% \subsubsection{Disclosure of Interests.}
% \pending{competing interests.}
% AUTHOR TODO: supply the applicable ethics or exemption statement.

\bibliographystyle{splncs04}
\bibliography{references}
\end{document}